\documentclass[nohyperref]{article}

\usepackage{microtype}
\usepackage{graphicx}
\usepackage{subcaption}
\usepackage{booktabs}

\usepackage{hyperref}

\newcommand{\acronym}{\text{LMD}}

\usepackage[accepted]{icml2023}

\usepackage{amsmath}
\usepackage{amssymb}
\usepackage{mathtools}
\usepackage{amsthm}

\usepackage[capitalize,noabbrev]{cleveref}

\theoremstyle{plain}

\theoremstyle{definition}

\theoremstyle{remark}

\usepackage[textsize=tiny]{todonotes}

\icmltitlerunning{Unsupervised Out-of-Distribution Detection with Diffusion Inpainting}

\begin{document}

\twocolumn[
\icmltitle{Unsupervised Out-of-Distribution Detection with Diffusion Inpainting}

\icmlsetsymbol{equal}{*}

\begin{icmlauthorlist}
\icmlauthor{Zhenzhen Liu}{equal,yyy}
\icmlauthor{Jin Peng Zhou}{equal,yyy}
\icmlauthor{Yufan Wang}{yyy}
\icmlauthor{Kilian Q. Weinberger}{yyy}
\end{icmlauthorlist}

\icmlaffiliation{yyy}{Department of Computer Science, Cornell University, Ithaca, New York, USA}

\icmlcorrespondingauthor{Zhenzhen Liu}{zl535@cornell.edu}
\icmlcorrespondingauthor{Jin Peng Zhou}{jz563@cornell.edu}

\icmlkeywords{Machine Learning, ICML}

\vskip 0.3in
]

\printAffiliationsAndNotice{\icmlEqualContribution} %

\begin{abstract}
Unsupervised out-of-distribution detection (OOD) seeks to identify out-of-domain data by learning only from unlabeled in-domain data. 
We present a novel approach for this task -- Lift, Map, Detect (LMD) -- that leverages recent advancement in diffusion models. Diffusion models are one type of generative models. At their core, they learn an iterative denoising process that gradually maps a noisy image closer to their training manifolds. 
\acronym{} leverages this intuition for OOD detection. Specifically, \acronym{} lifts an image off its original manifold by corrupting it, and maps it towards the in-domain manifold with a diffusion model. 
For an out-of-domain image, the mapped image would have a large distance away from its original manifold, and \acronym{} would identify it as OOD accordingly.  We show through extensive experiments that \acronym{} achieves competitive performance across a broad variety of datasets. Code can be found at \url{https://github.com/zhenzhel/lift_map_detect}.

\end{abstract}

\section{Introduction}

\begin{figure}[t]
    \centering
    \includegraphics[width=\linewidth,trim={0 1cm 0 0cm},clip]{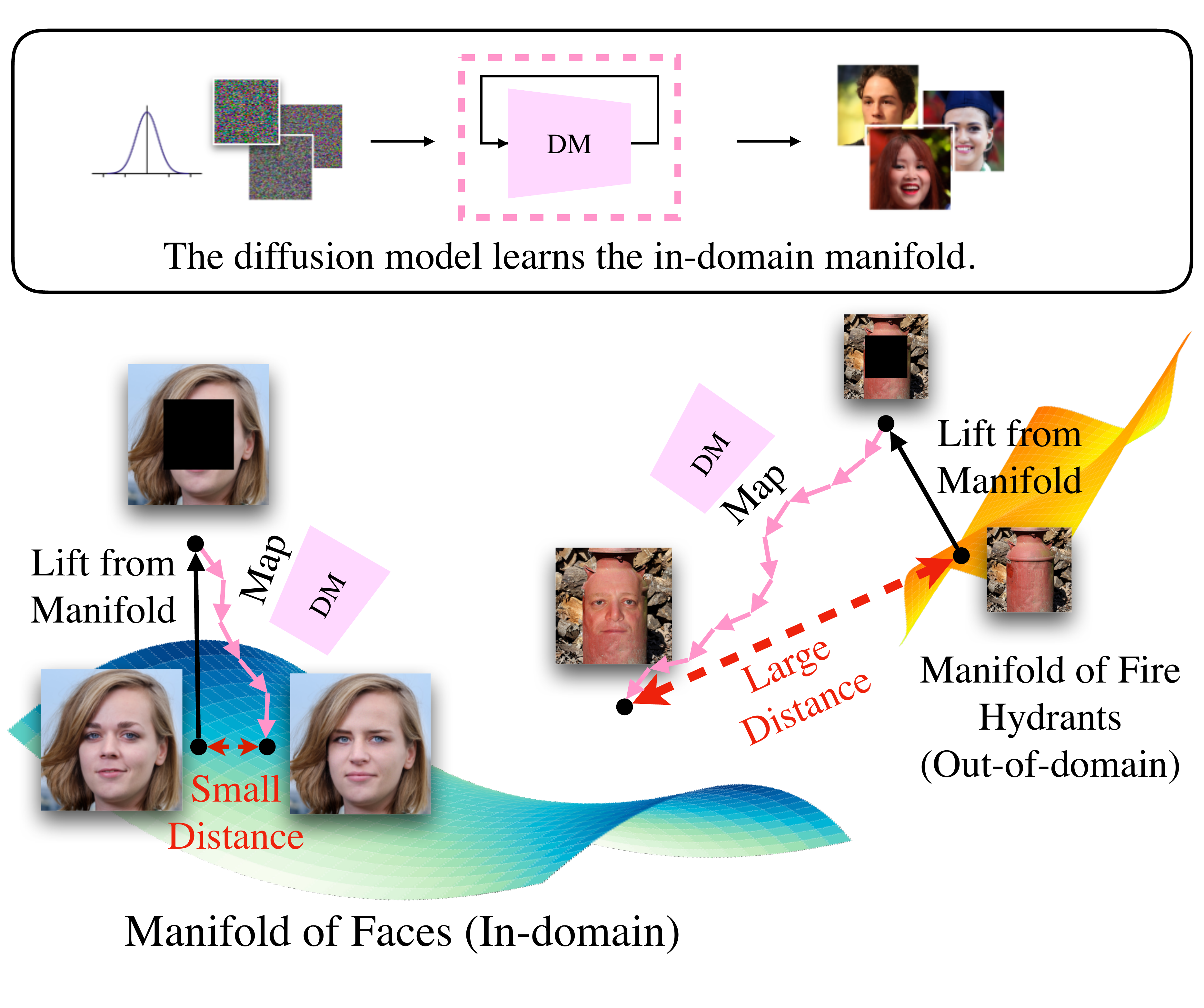}
    \caption{The pictorial intuition behind \acronym{} for OOD detection. A diffusion model learns a mapping to the in-domain manifold. \acronym{} lifts an image off its manifold by masking, and uses the diffusion model to move it towards the in-domain manifold. An in-domain image would have a much smaller distance between the original and mapped locations than its out-of-domain counterparts.}
    \label{fig:manifold}
\end{figure}

Out-of-distribution (OOD) detection seeks to classify whether a data point belongs to a particular domain. It is especially important, because machine learning models typically assume that test-time samples are drawn from the same distribution as the training data. If the test data do not follow the training distribution, they can inadvertently produce non-sensical results. The increased use of machine learning models in high-stake areas, such as medicine~\cite{hamet2017medicine} and criminal justice~\cite{rigano2019criminal}, amplifies the importance of OOD detection. For example, if a doctor mistakenly inputs a chest X-ray into a brain tumor detector, the model would likely still return a prediction -- which would be meaningless and possibly misleading. 

Previous researches have studied OOD detection under different settings: supervised and unsupervised. 
    Within the supervised setup, the supervision can originate from different sources. In the most informed setting, one assumes access to representative out-of-domain samples. These allow one to train an OOD detector as a classifier distinguishing in-domain from out-of-domain data, and achieve high performance~\cite{hendrycks2018deep, ruff2019deep} -- as long as the out-of-domain data do not deviate from the assumed out-of-domain distribution. In many practical applications, however, such knowledge is unattainable. In fact, out-of-domain data can be highly diverse and unpredictable. A significantly more relaxed assumption is to only require access to an in-domain classifier or class labels. Under this setting, methods such as ~\citet{hendrycks2016baseline, liang2017enhancing,lee2018simple, huang2021importance, wang2022vim} have achieved competitive performance. Although less informed, this setting relies on two implicit assumptions: the in-domain data have well-defined classes, and there are sufficiently plenty data with class annotations. In practice, these assumptions often cannot be met. Unlabeled data do not require the expensive human annotation, and thus are often readily available in large quantity. Ideally, one would like to build an OOD detector that only requires unlabeled in-domain data during training. 

Recently, a class of generative models -- the diffusion models (DM)~\cite{ho2020denoising, song2020score} -- have gained increasing popularity. 
DMs formulate two processes: The forward process converts an image to a sample drawn from a noise distribution by iteratively adding noise to its pixels; the backward process maps a noise image towards a specific image manifold by iteratively removing noise from the image. A dedicated neural network is trained to perform the denoising steps in the backward process. 

In this paper, we argue that we can leverage the property that the diffusion model learns a mapping to a manifold, and turn it into a strong unsupervised OOD detector. Intuitively, if we lift an image from its manifold, then the lifted image can be mapped back to its original vicinity with a diffusion model trained over the same manifold. If instead the diffusion model is trained over a different manifold, it would attempt to map the lifted image towards its own training manifold, causing a large distance between the original and mapped images. Thus, we can detect out-of-domain images based on such distance. 

To this end, we propose a novel unsupervised OOD detection approach called \textbf{L}ift, \textbf{M}ap, \textbf{D}etect (\acronym{}) that captures the above intuition. We can \textbf{lift} an image from its original manifold by corrupting it. For example, a face image masked in the center clearly does not belong to the face manifold anymore. As shown by \citet{song2020score, lugmayr2022repaint}, the diffusion model can impute the missing regions of an image with visually plausible content, a process commonly referred as inpainting, without retraining. Thus, we can \textbf{map} the lifted image by inpainting with a diffusion model trained over the in-domain data. We can then use a standard image similarity metric to measure the distance between the original and mapped images, and \textbf{detect} an out-of-domain image when we observe a large distance. Figure \ref{fig:manifold} illustrates an example: A diffusion model trained on face images maps a lifted in-domain face image closer to the original image than an out-of-domain fire hydrant counterpart. 

To summarize our contributions:
 \begin{enumerate}
     \item We propose a novel approach \acronym{} for unsupervised OOD detection, which directly leverages the diffusion model's manifold mapping ability without retraining. We introduce design choices that improve the separability between in-domain and out-of-domain data.
     \item We show that \acronym{} is versatile through experiments on datasets with different coloring, variability and resolution.
     \item We provide qualitative visualization and quantitative ablation results that verify the basis of our approach and our design choices.
 \end{enumerate}

\section{Background}

\textbf{Unsupervised OOD Detection.} We formalize the unsupervised OOD detection task as follows: Given a distribution of interest $\mathcal{D}$, one would like to build a detector that decides whether a data point $\mathbf{x}$ is drawn from $\mathcal{D}$. The detector is only built upon unlabeled in-distribution samples $\mathbf{x_1}, \cdots, \mathbf{x_n} \sim \mathcal{D}$. Given a test data point $\mathbf{x}$, the detector outputs an OOD score $s(\mathbf{x})$, where a higher $s(\mathbf{x})$ signifies that $\mathbf{x}$ is more likely \emph{not} to be sampled from $\mathcal{D}$.

Existing works can be roughly divided into three categories: likelihood-based, reconstruction-based, and feature-based. Likelihood-based approaches date back to~\citet{bishop1994novelty}. At a high level, one fits the in-domain distribution with a model, and evaluates the likelihood of the test data under the model. Recent approaches often employ a deep generative model that supports likelihood computation, such as PixelCNN++~\cite{salimans2017pixelcnn++} or Glow~\cite{kingma2018glow}. However, several works \cite{choi2018waic, nalisnick2018deep, kirichenko2020normalizing} have found that generative models sometimes assign higher likelihood to out-of-domain data. 

This issue can be alleviated in various ways. One line of work adopts a likelihood ratio approach: \citet{ren2019likelihood} trains a semantic model and a background model, and takes the ratio of the likelihoods from the two models. ~\citet{serra2019input} observes a negative correlation between an image's complexity and its likelihood, and adjusts the likelihood by the compression size. ~\citet{xiao2020likelihoodregret} optimizes the model configuration to maximize a test image's likelihood, and measures the amount of likelihood improvement. Another line of work adopts a typicality test approach~\cite{nalisnick2019detecting, morningstar2021density,bergamin2022agnostic}. They examine the distribution of in-domain likelihood or other model statistics, and evaluate the typicality of the test data model statistics through hypothesis testing or density estimation. Lastly, several works~\cite{maaloe2019biva,kirichenko2020normalizing} seek to improve the design choices of generative models.

Reconstruction-based approaches evaluate how well a data point can be reconstructed by a model learned over the in-domain data. Our approach \acronym{} falls into this category. Within this line of work, \citet{sakurada2014anomaly, xia2015learning, zhou2017anomaly, zong2018deep} encode and decode data with autoencoders. \citet{schlegl2017unsupervised, li2018anomaly} perform GAN~\cite{goodfellow2014gan} inversion for a data point, and evaluate its reconstruction error and discriminator confidence under the inverted latent variable. Additionally, concurrent to our work, \citet{graham2022denoising} leverages diffusion models to reconstruct images at varied diffusion steps, while we mask and inpaint an image repeatedly with fixed steps. The two approaches are complementary to each other. 

Feature-based approaches featurize data in an unsupervised manner, and fit a simple OOD detector like a Gaussian Mixture Model over the in-domain features. \citet{denouden2018improving} leverages the latent variables of an autoencoder, and evaluates the Mahalanobis distance in the latent space along with the data reconstruction error. \citet{ahmadian2021likelihood} extracts low-level features from the encoder of an invertible generative model. \citet{hendrycks2019using, bergman2020classification, tack2020csi, sehwag2021ssd} learn a representation over the in-domain data through self-supervised training; \citet{xiao2021we} further shows that one can instead use a strong pretrained feature extractor while maintaining comparable performance. 

\textbf{Diffusion Models.} In this section, we provide a brief overview of the diffusion models (DM). It is a type of generative model that learns the distribution of its training data. DM formulates a forward process of corrupting data by adding noise to them, commonly referred as diffusion. It learns the reverse process of gradually producing a less noisy sample, commonly referred as denoising. One classic formulation of DM is called Denoising Diffusion Probabilistic Models (DDPMs)~\cite{sohl2015deep,ho2020denoising}. Specifically, starting from a data sample $x_0$, each step $t = 1, 2, \cdots, T$ of the diffusion process injects Gaussian noise given by
\begin{equation}
    q(x_t|x_{t-1}) = \mathcal{N}(x_t;\sqrt{1-\beta_t}x_t, \beta_t\mathbf{I})
\end{equation}
where $\beta_t$ follows a fixed variance schedule. The DDPM with a prior distribution $x_T \sim \mathcal{N}(0, 1)$ learns the denoising process given by
\begin{equation}
    p_\theta(x_{t-1}|x_t) = \mathcal{N}(x_{t-1};\mu_\theta(x_t, t), \Sigma_\theta(x_t, t))
\end{equation}
where both $\mu_\theta(x_t, t)$ and $\Sigma_\theta(x_t, t)$ are learned by a neural network parametrized by $\theta$.
Note that other formulations of DMs, such as score-based generative models~\cite{song2019generative} and stochastic differential equations~\cite{song2020score}, also support diffusion and denoising processes. Since \acronym{} is agnostic to different formulations of DMs, we refer the reader to~\citet{yang2022dmsurvey} for a more detailed mathematical description of the other formulations.

\section{Lift, Map, Detect}\label{sec:methods}

\begin{figure*}[t]
\centering
\includegraphics[width=0.9\linewidth]{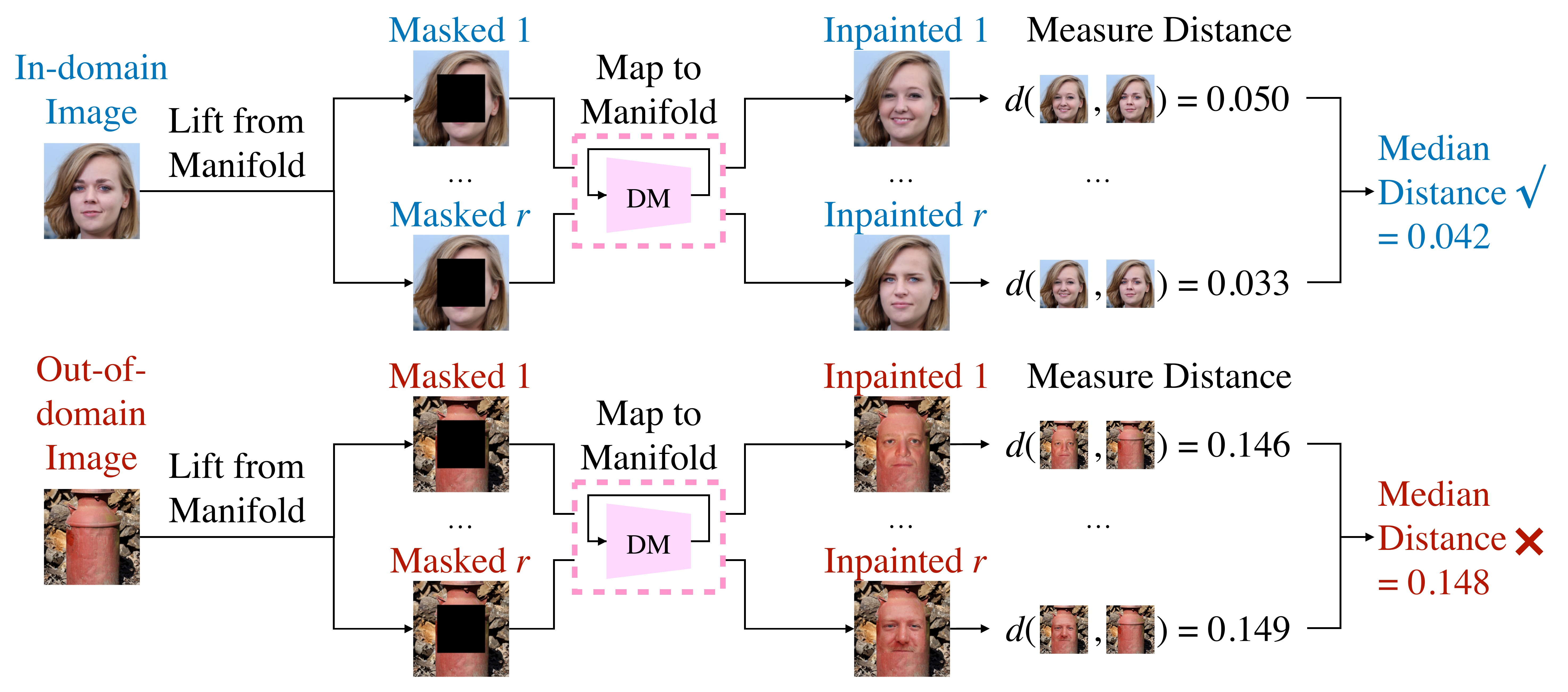}
    \caption{High-level workflow of \acronym{}. \acronym{} employs a diffusion model learned over the in-domain manifold. It first repeatedly lifts an image from its manifold by masking it, and maps it towards the diffusion model's training manifold by inpainting. Then, it inspects the median distance between the original image and each mapped image to detect out-of-domain images. As out-of-domain images cannot be mapped back to their own manifolds, they have larger distances.}
\label{fig:overview}
\end{figure*}

The intuition behind our algorithm, Lift, Map, Detect (\acronym{}), is illustrated in Figure \ref{fig:manifold}. In a nutshell, we employ a diffusion model learned over the in-domain data, which provides a mapping towards the underlying in-domain image manifold. To test whether an image is in-domain or out-of-domain, we lift the image off its original manifold through corruption, and map the lifted image to the in-domain manifold with the trained DM. If the original image is in-domain, it is mapped back to its manifold, near its original location. If it is out-of-domain, the image is mapped to a different manifold, likely leaving a large distance between the original image and the mapped image. Figure \ref{fig:overview} shows the high-level workflow of \acronym{}. Algorithm \ref{alg:mid} summarizes the key steps of \acronym{} in pseudocode.

\textbf{Lifting Images.} To lift an image off its manifold, we need to corrupt the image so that it no longer appears to be from its original manifold. Concretely, we apply a mask to the image so that part of it is completely removed.  Since various mask patterns and sizes can be used, masking provides a direct and flexible way to lift the image from the manifold. For example, it is intuitive to see that the larger the mask is, the further away the image is lifted from the manifold.

\textbf{Mapping the Lifted Images.} Since diffusion models (DM) can perform inpainting without retraining~\cite{song2020score,lugmayr2022repaint} (see Algorithm \ref{alg:inpainting}), we naturally employ a DM and use inpainting to map the lifted images. Specifically, we employ a DM parametrized by $\theta_{in}$ that is trained on the in-domain data. This DM can model the in-domain distribution well enough to map a lifted in-domain image back to its original vicinity. Meanwhile, the DM should have almost no knowledge about the out-of-domain manifold. Thus, it naturally maps a lifted out-of-domain image towards the DM's training manifold, which is the in-domain manifold. This phenomenon leads to a larger distance between the original and mapped images for out-of-domain images than the in-domain ones. For ease of reference, we also refer these mapped images as reconstructed images or simply reconstructions.

\begin{algorithm}[htb]
   \caption{Inpaint}
   \label{alg:inpainting}
\begin{algorithmic}
   \STATE {\bfseries Input:} original image $x_{orig}$, binary mask $M$ where 0 indicates region to be inpainted, diffusion model $\theta$
   \STATE {\bfseries Output:} inpainted image $x_{inp}$
   \FOR{$t=T$ {\bfseries to} $1$}
   \IF{$t == T$}
   \STATE $x_{inp}$ $\gets$ sample from noise distribution
   \ENDIF
   \STATE $x_{orig}'$ $\gets$ \texttt{diffuse($x_{orig};\: \theta$)} to step $t - 1$
   \STATE $x_{inp}$ $\gets$ \texttt{denoise($x_{inp};\:\theta$)} to step $t - 1$
   \STATE $x_{inp}$ $\gets$ $x_{orig}' \cdot M + x_{inp} \cdot (1 - M)$
   \ENDFOR
   \STATE {\bfseries return} $x_{inp}$
\end{algorithmic}
\end{algorithm}

\textbf{Reconstruction Distance Metric.}  We adopt the Learned Perceptual Image Patch Similarity (LPIPS)~\cite{zhang2018lpips} metric, a standard and strong metric that captures the perceptual difference between images. Since LPIPS assigns higher values to more dissimilar images, we compute the LPIPS between original and reconstructed images, and use it directly as the OOD score. We perform detailed ablation on different reconstruction distance metrics in Section \ref{sec:ablation}. 
 
It is worth noting that mapping lifted images is the most crucial component. The hypothesis -- in-domain reconstructions are closer to their original images than the out-of-domain ones -- ensures the effectiveness of \acronym{}. With this in mind, we now discuss two simple and yet effective ways that can further improve detection performance consistently: multiple reconstructions and novel masking strategy.

\textbf{Multiple Reconstructions.} The DM inpainting process inherently involves multiple sampling steps. Occasionally, due to randomness, DM could provide dissimilar reconstructions for in-domain data, or similar reconstructions for out-of-domain data. This could make the reconstruction distance of the in-domain and out-of-domain images less separable, and hence lead to suboptimal OOD detection performance. To reduce the randomness, we perform multiple lifting and mapping attempts for each image. We calculate the OOD score from each attempt, and take the median\footnote{In our preliminary experiments, we find that median works better than other simple aggregation methods such as mean.} OOD score as the final OOD score for an image. As shown in Section~\ref{sec:main_results}, the simple median aggregation already provides strong performance. For further improvement, one may use a parameterized model to learn the distribution of the reconstruction distance across multiple attempts. We leave this to future work.

\begin{algorithm}[htb]
   \caption{Lift, Map, Detect (\acronym{})}
   \label{alg:mid}
\begin{algorithmic}
   \STATE {\bfseries Input:} test image $x$, in-domain diffusion model $\theta_{in}$
   \STATE {\bfseries Output:} OOD score of test image $x$
   \FOR{$i=1$ {\bfseries to} $r$}
   \STATE $M_{i}$ $\gets$ \texttt{Get\_Mask($i$)}
   \STATE $x_{i}'$ $\gets$ \texttt{Inpaint($x, M_{i}, \theta_{in}$)}
   \STATE $d_{i}$ $\gets$ \texttt{Distance($x, x_{i}'$)}
   \ENDFOR
   \STATE {\bfseries return} \texttt{Aggregate($d_1, \ldots, d_r$)}
\end{algorithmic}
\end{algorithm}

\begin{figure}[h]
\centering
\includegraphics[width=\linewidth]{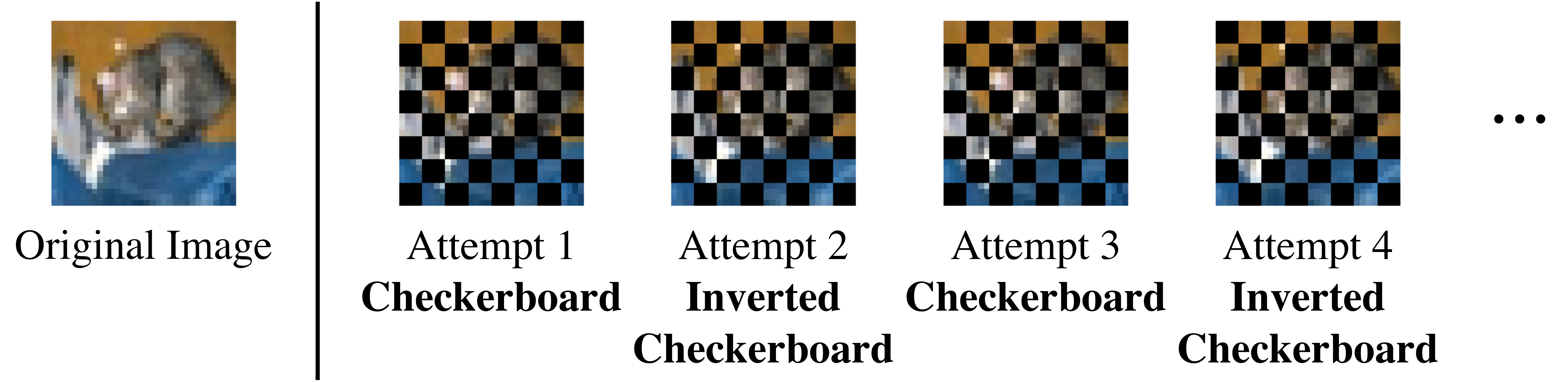}
    \caption{The alternating checkerboard mask pattern. We invert regions that are masked and unmasked at each reconstruction attempt. The example in the figure is $8\times 8$.}
\label{fig:flip}
\end{figure}

\textbf{Novel Masking Strategy.} 
The extent to which we mask an image is crucial to the detection performance. If the size of the mask is too large (or too small), the reconstruction distance for both in-domain and out-of-domain images would be very large (or very small). Indeed, if the mask covers the entire image, the reconstruction will be independent of the original image. In this case, if the in-domain manifold contains diverse images, an in-domain reconstruction can be far from its original image despite still being on the same manifold. Therefore, a suitable masking strategy should leave enough context to allow in-domain reconstructions to be similar to the original ones. To this end, we propose to use a checkerboard mask pattern. It divides an image into an $N \times N$ grid of image patches independent of the image size, and masks out half of the patches similar to a checkerboard. When multiple reconstruction attempts are performed, we also invert the masked and unmasked regions at each attempt. We call this masking strategy \textit{alternating checkerboard $N \times N$} (see Figure \ref{fig:flip}). Alternating checkerboard ensures all regions of the image to be masked with just two attempts. This avoids situations in which the distinguishing features of an out-of-domain image is never masked. \acronym{} by default sets $N = 8$; ablation study on different mask choices can be found in Table~\ref{tab:masks}. 

\section{Experiments}

\subsection{Experiment Settings}
\textbf{Evaluation Metric.} \acronym{} outputs an OOD score for each input, so in practice we need to apply a threshold to binarize the decision. In the experiments, we follow~\citet{hendrycks2016baseline, ren2019likelihood,xiao2021we}, and use the area under Receiver Operating Characteristic curve (ROC-AUC) as our quantitative evaluation metric. 

\textbf{Baselines.} We compare our methods with seven existing baselines: Likelihood \textbf{(Likelihood)} ~\cite{bishop1994novelty}, Input Complexity \textbf{(IC)}~\cite{serra2019input}, Likelihood Regret \textbf{(LR)}~\cite{xiao2020likelihoodregret}, Pretrained Feature Extractor + Mahalanobis Distance \textbf{(Pretrained)}~\cite{xiao2021we}, Reconstruction with Autoencoder and Mean Squared Error loss \textbf{(AE-MSE)}, AutoMahalanobis \textbf{(AE-MH)}~\cite{denouden2018improving} and AnoGAN \textbf{(AnoGAN)}~\cite{schlegl2017unsupervised}.  Likelihood is obtained from the DM using the implementation from~\citet{song2020score}\footnote{https://github.com/yang-song/score\_sde\_pytorch}. For both Input Complexity and Likelihood Regret, we adapt the official GitHub repository of Likelihood Regret\footnote{https://github.com/XavierXiao/Likelihood-Regret}. Specifically, to compute the Input Complexity, we use the likelihood calculated from the DM for a fair comparison, and convert the compression size to bits per dimension; we use the PNG compressor, because it yields the best performance among all available compressors in the GitHub repository. Pretrained Feature Extractor + Mahalanobis Distance is implemented by ourselves, as there is no existing publicly available implementation to our best knowledge.

\textbf{Datasets.} We perform OOD detection pairwise among CIFAR10~\cite{Krizhevsky09learningmultiple}, CIFAR100~\cite{Krizhevsky09learningmultiple} and SVHN~\cite{Netzer2011}, and pairwise among MNIST~\cite{lecun2010mnist}, KMNIST~\cite{clanuwat2018deep} and FashionMNIST~\cite{DBLP:journals/corr/abs-1708-07747}. For \acronym{} and all the baselines, we use the training set of the in-domain dataset to train the model if needed, and evaluate the performance on the full test set of in-domain and out-of-domain datasets. Additionally, to demonstrate our performance on higher resolution images, we show qualitative results on CelebA-HQ~\cite{DBLP:journals/corr/abs-1710-10196} as in-domain and ImageNet~\cite{ILSVRC15} as out-of-domain. 

\begin{table*}[h]
\caption{ROC-AUC performance of \acronym{} against various baselines on 12 pairs of datasets. Higher is better. We use the same configuration for \acronym{} across all datasets: Alternating checkerboard mask $8 \times 8$, distance metric LPIPS, and 10 reconstructions per image. \acronym{} consistently demonstrates strong performance and attains the highest average ROC-AUC. }
\label{tab:main}
\vskip 0.15in
\begin{center}
\begin{small}
\begin{sc}
  \resizebox{0.9\linewidth}{!}{%
\begin{tabular}{lccccccccc}
\toprule
ID & OOD  & Likelihood & IC & LR & Pretrained & AE-MSE & AE-MH & AnoGAN & \acronym{} \\
\midrule
CIFAR10    & CIFAR100 & 0.520 & 0.568 & 0.546 & \textbf{0.806} & 0.510 & 0.488 & 0.518 & 0.607 \\
         &    SVHN  & 0.180 & 0.870 & 0.904 & 0.888 & 0.025 & 0.073 & 0.120 & \textbf{0.992} \\
 \midrule
CIFAR100    & CIFAR10 & 0.495 & 0.468  & 0.484 & 0.543 & 0.509 & 0.486 & 0.510 & \textbf{0.568}\\
         &    SVHN & 0.193 & 0.792 &  0.896 & 0.776 & 0.027 & 0.122 & 0.131 & \textbf{0.985} \\
\midrule
SVHN    & CIFAR10 & 0.974 & 0.973 & 0.805 & \textbf{0.999} & 0.981	& 0.966 & 0.967 & 0.914\\
         &    CIFAR100 & 0.970 & 0.976 & 0.821 & \textbf{0.999} & 0.980 & 0.966 & 0.962 & 0.876\\
\midrule
MNIST    & KMNIST & 0.948 & 0.903 & 0.999 & 0.887 & 0.999 & \textbf{1.000} & 0.933 & 0.984\\
         &    FashionMNIST & 0.997 & \textbf{1.000} & 0.999 & 0.999 & \textbf{1.000} & \textbf{1.000} & 0.992 & 0.999\\
\midrule
KMNIST    & MNIST & 0.152 & 0.951 & 0.431 & 0.582 & 0.102 & 0.217 & 0.317 & \textbf{0.978}\\
         &    FashionMNIST & 0.833 & \textbf{0.999} & 0.557 & 0.993 & 0.896 & 0.868 & 0.701 & 0.993  \\
\midrule
FashionMNIST    & MNIST & 0.172 & 0.912 &  0.971 & 0.647 & 0.804 & 0.969 & 0.835 & \textbf{0.992}\\
         &    KMNIST  & 0.542 & 0.584 & 0.994 & 0.730 & 0.976 & \textbf{0.996} & 0.912 & 0.990\\
\midrule
Average & & 0.581 & 0.833 & 0.783 & 0.821 & 0.651 & 0.679 & 0.658 & \textbf{0.907}\\
\bottomrule
\end{tabular}
}
\end{sc}
\end{small}
\end{center}
\vskip -0.1in
\end{table*}

\subsection{Implementation Details of \acronym{}}
We adapt the diffusion model implementation from~\citet{song2020score}. For experiments in Table~\ref{tab:main}, we use \citet{song2020score}'s pretrained checkpoint for CIFAR10, and we train DMs on the training set of the in-domain dataset for all the other datasets. We evaluate the OOD scores of the full in-domain and out-of-domain test sets. The inpainting reconstruction is repeated 10 times with alternating checkerboard $8\times 8$ masks (Figure~\ref{fig:flip}). For CelebA-HQ vs. ImageNet, we observe that CelebA-HQ does not have a train/test set split, and its pretrained checkpoint is trained over the full dataset. Thus, to avoid potential memorization issues, we use the pretrained FFHQ~\cite{karras2019style} checkpoint instead. We randomly sample a subset of size 100 from each dataset, and standardize all images to $256 \times 256$. We explore three mask choices: checkerboard $4\times 4$, checkerboard $8 \times 8$, and a square-centered mask. We reconstruct each image only {\em once}. We use LPIPS as the reconstruction distance metric to calculate the OOD score for all the experiments.

\subsection{Experimental Results}\label{sec:main_results}
Table~\ref{tab:main} shows the performance of \acronym{} and the baselines on various pairs of datasets. \acronym{} achieves the highest performance on five pairs, with a maximum improvement of 10\% (CIFAR100 vs. SVHN). \acronym{} also achieves competitive performance on several other pairs, and attains the highest average ROC-AUC. This shows that \acronym{} is consistent and versatile. We observe that the performance of the baselines are competitive on some pairs but limited on the others. 

\begin{figure*}[t]%
\centering
\includegraphics[width=0.8\linewidth,trim={0 1cm 0 1cm},clip]{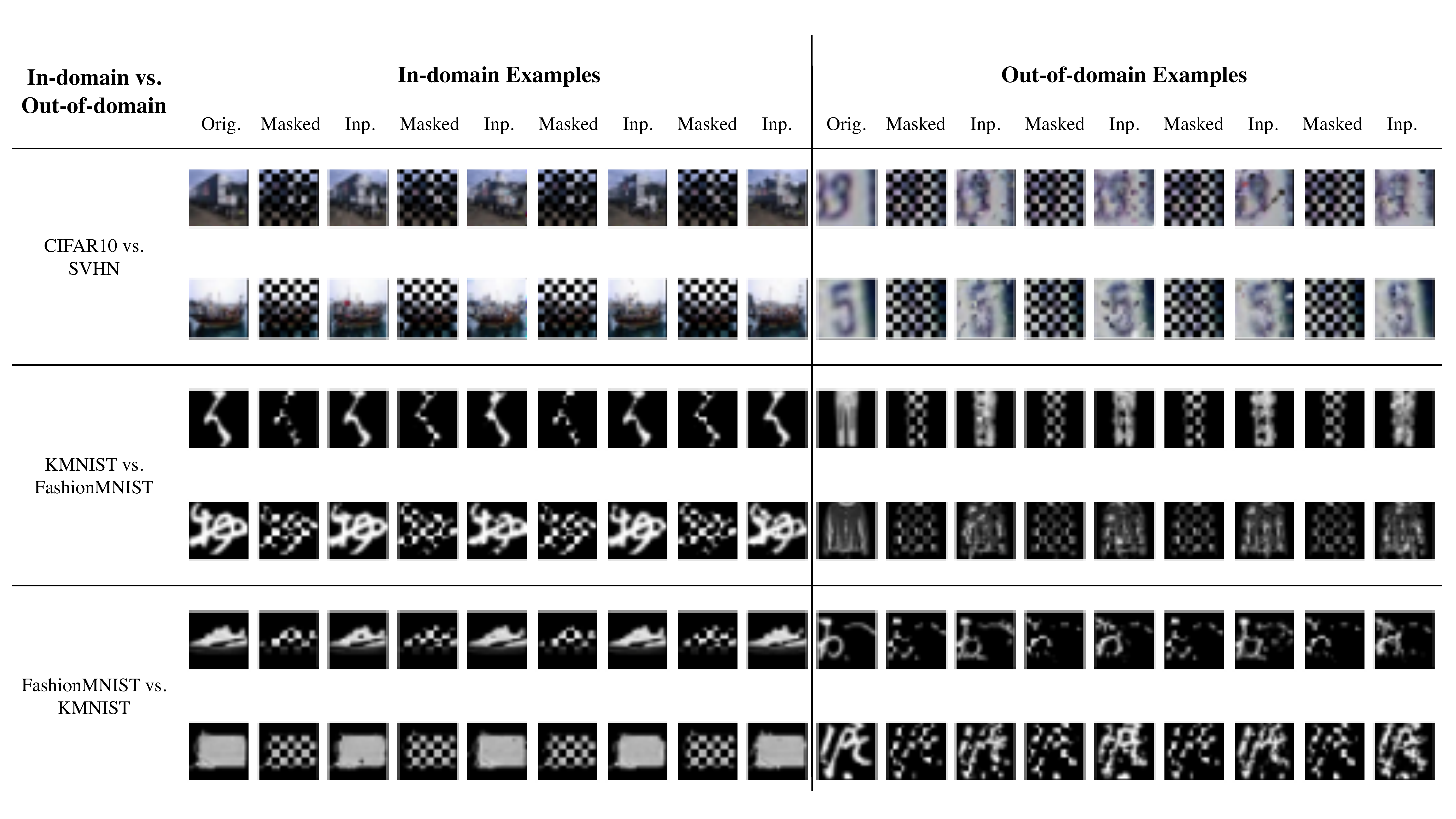}
    \caption{Reconstruction examples from three dataset pairs. ``Orig." stands for the original image; ``Inp." stands for the inpainted image. In general, the in-domain reconstructions are close to their original images, while the out-of-domain reconstructions are noisy and different from the original ones.}
\label{fig:small}
\end{figure*}

Figure~\ref{fig:small} shows examples of the original, masked and inpainted images for three pairs. We show four reconstruction examples for each image, two with checkerboard mask and two with inverted checkerboard mask. The diffusion models reconstruct the in-domain images relatively accurately, while introducing a lot of artifacts in the out-of-domain inpaintings. For example, when SVHN is out-of-domain, the noise almost overwhelms the signals in the inpaintings. 

\begin{figure*}[t]%
\centering
\includegraphics[width=\linewidth]{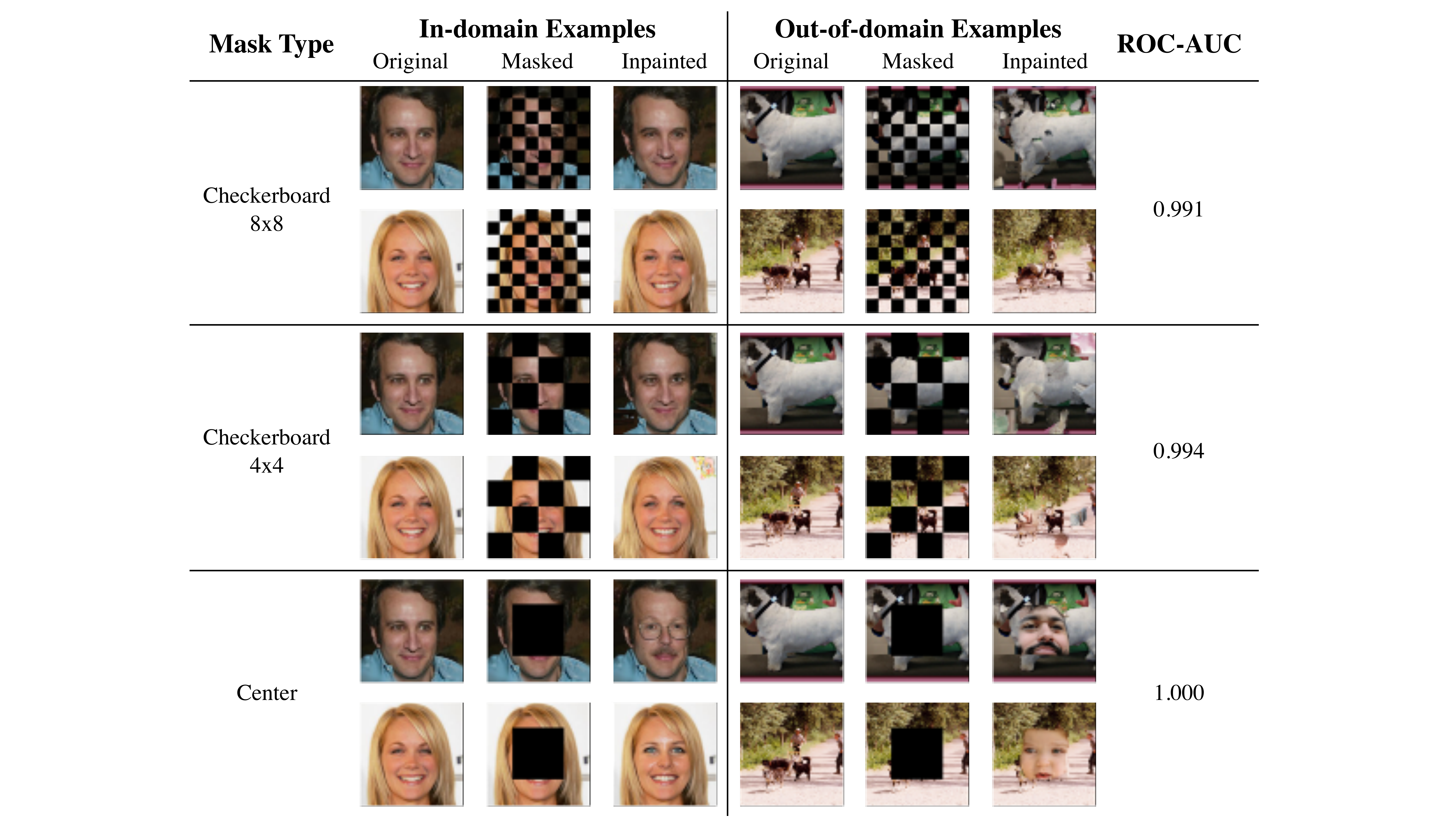}
    \caption{Reconstruction examples from CelebA-HQ (in-domain) and ImageNet (out-of-domain) using different masks. For out-of-domain inpaintings, the checkerboard masks introduce locally incoherent artifacts, while the center mask introduces face-like artifacts. This makes the out-of-domain images highly distinguishable.}
\label{fig:celeba}
\end{figure*}

Figure~\ref{fig:celeba} shows the qualitative results and the ROC-AUC for CelebA-HQ vs. ImageNet. Checkerboard $8\times 8$ performs competitively, achieving an ROC-AUC of 0.991 without any repeated reconstructions. Visually, the in-domain inpaintings look almost identical to the original images, while the out-of-domain inpaintings are locally incoherent. In this specific setting, checkerboard $4\times 4$ and center masks yield slightly better performance. This is probably because faces are highly structured and provide a strong inductive bias. Thus, with larger contiguous masked regions, the DM can still produce reasonably authentic reconstructions for the in-domain images, while being able to introduce more obvious artifacts for the out-of-domain images. Consequently, the reconstruction qualities of the two domains are more distinguishable. More discussion on mask choices can be found in Section~\ref{sec:ablation}.

\begin{figure}[h]%
\centering
\includegraphics[width=\linewidth,trim={0 11cm 0 11cm},clip]{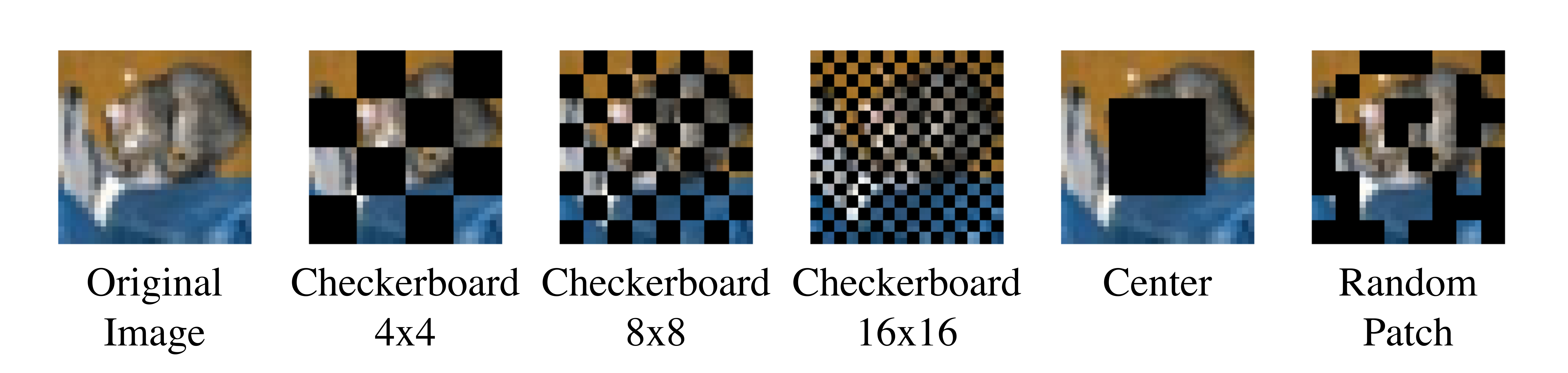}
    \caption{Visualization of the masks used in the mask ablation. For the random patch mask, this figure only shows one example; we sample a different pattern at each reconstruction attempt.}
\label{fig:masks}
\end{figure}

\begin{table*}[h]
\caption{ROC-AUC performance on three dataset pairs with different mask types. Alternating checkerboard $8\times8$ shows strong and consistent performance.}
\begin{center}
\begin{sc}
\vskip 0.15in
  \resizebox{0.9\linewidth}{!}{
  \begin{tabular}{lccc}
    \toprule
    Mask Type & CIFAR10 vs. CIFAR100 & CIFAR10 vs. SVHN & MNIST vs. KMNIST\\
    \midrule
    Alternating Checkerboard $4\times4$ & 0.594 & 0.987 & 0.923\\
    Alternating Checkerboard $8\times8$& \textbf{0.607} & \textbf{0.992} & 0.984\\
    Alternating Checkerboard $16\times16$ & 0.597 & 0.981 & \textbf{0.997}\\
    Fixed Checkerboard $8\times8$ & 0.601 & 0.990 & 0.974 \\
    Center & 0.570 & 0.978 & 0.479\\
    Random Patch & 0.591 & 0.990 & 0.912\\
    \bottomrule
  \end{tabular}
  }
\end{sc}
\end{center}
\label{tab:masks}
\end{table*}

\subsection{Ablation}\label{sec:ablation}%

{\bf Effects of Mask Choices.} Table~\ref{tab:masks} shows ablation results on different types of mask patterns (see Figure \ref{fig:masks}). Specifically, we examine the following patterns: alternating checkerboard $4 \times 4$ and $16 \times 16$, a fixed non-alternating $8 \times 8$ checkerboard, a square centered mask covering one-fourth of an image \textbf{(center)}, and a random patch mask covering 50\% of an $8 \times 8$ patch grid \textbf{(random patch)} introduced in~\citet{xie2022simmim}\footnote{https://github.com/microsoft/SimMIM}.

Alternating checkerboard $8 \times 8$ performs consistently across the three datasets, while other patterns have fluctuation in their performance. Not surprisingly, the center mask exhibits very poor performance (0.444) on MNIST vs. KMNIST, as it removes too much information from the images. Alternating checkerboard $4 \times 4$ also underperforms on MNIST vs. KMNIST. This suggests that if the masked patches are too large, both in-domain and out-of-domain reconstructions may be dissimilar from the original images. Fixed checkerboard $8 \times 8$ performs only slightly worse than its alternating counterpart, usually with a performance drop of less than 0.01. This may be because for these datasets, the distinguishing features of the in-domain and out-of-domain images exist in many patches. Thus, they can already be captured well enough by the fixed $8 \times 8$ mask. Nevertheless, the alternating checkerboard pattern should still be preferred, since it can mask the entire image across multiple reconstruction attempts. 

{\bf Effects of Reconstruction Distance Metrics.} \acronym{} needs to assess the reconstruction distance of the DM's inpaintings, so we explore three off-the-shelf reconstruction distance metrics: Mean Squared Error (MSE), Structural Similarity Index Measure (SSIM)~\cite{wang2003ssim}, and Learned Perceptual Image Patch Similarity (LPIPS)~\cite{zhang2018lpips}. Additionally, \citet{xiao2021we} demonstrates strong performance using SimCLRv2~\cite{chen2020simclrv2} representations, so we experiment with a SimCLRv2-based error metric too. Specifically, we calculate the cosine distance between the SimCLRv2 representations of the original and reconstructed images, which we simply refer to as SimCLRv2. These four reconstruction distance metrics range from shallow reference based to deep feature based metrics.

\begin{table}[h]
\caption{ROC-AUC performance on three dataset pairs with different reconstruction distance metrics. LPIPS attains consistent and strong performance, while other metrics have fluctuation in their performance.}
\vskip 0.15in
\begin{center}
\begin{small}
\begin{sc}
\resizebox{\linewidth}{!}{
\begin{tabular}{lcccr}
\toprule
Recon. Metric & CIFAR10  & CIFAR10 & KMNIST \\
& vs. CIFAR100 & vs. SVHN & vs. MNIST\\
\midrule
MSE & 0.548 & 0.155 & 0.835\\
SSIM & 0.624 & 0.329 & 0.922\\
LPIPS & 0.607 & \textbf{0.992} & \textbf{0.978}\\
SimCLRv2 & \textbf{0.713} & 0.970 & 0.920\\
\bottomrule
\end{tabular}
}
\end{sc}
\end{small}
\end{center}
\vskip -0.1in
\label{tab:metrics}
\end{table}

We summarize the results of three dataset pairs in Table~\ref{tab:metrics}. LPIPS is competitive on all three dataset pairs, while MSE, SSIM, and SimCLRv2 fluctuate in their performance. Interestingly, on CIFAR10 vs. CIFAR100, SimCLRv2 outperforms other metrics significantly, with an improvement of 0.09. In Table \ref{tab:main}, \citet{xiao2021we} also outperforms all other methods on CIFAR10 vs. CIFAR100 using SimCLRv2 representations. This suggests that other metrics may be suitable for specific domains, and LPIPS serves as an effective default choice for the distance metric.

\begin{figure}[h]%
\centering
\begin{subfigure}{0.49\linewidth}
\centering
\includegraphics[width=\linewidth]{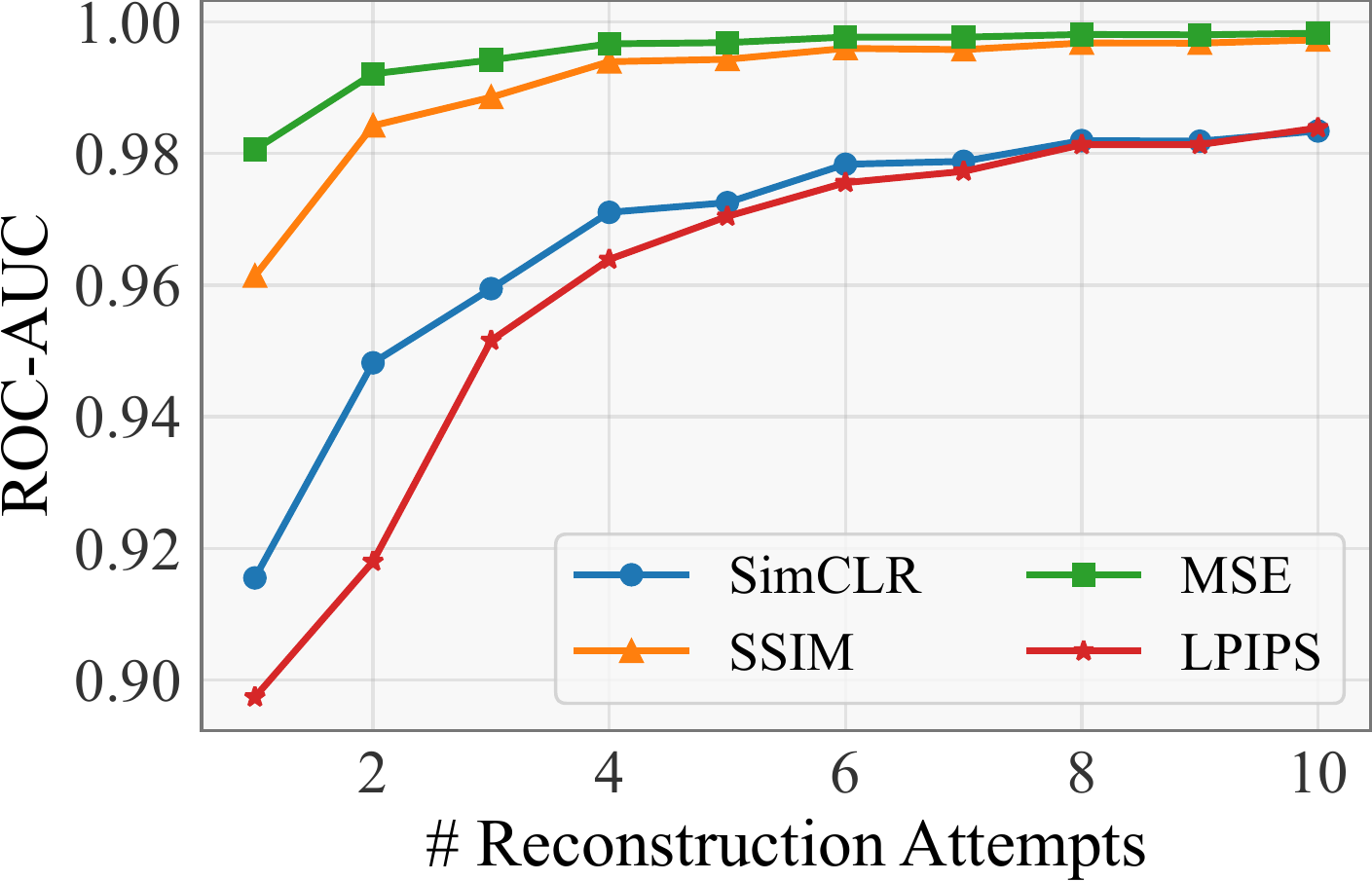}
\caption{MNIST vs. KMNIST}
\end{subfigure}\hfill
\begin{subfigure}{0.49\linewidth}
\centering
\includegraphics[width=\linewidth]{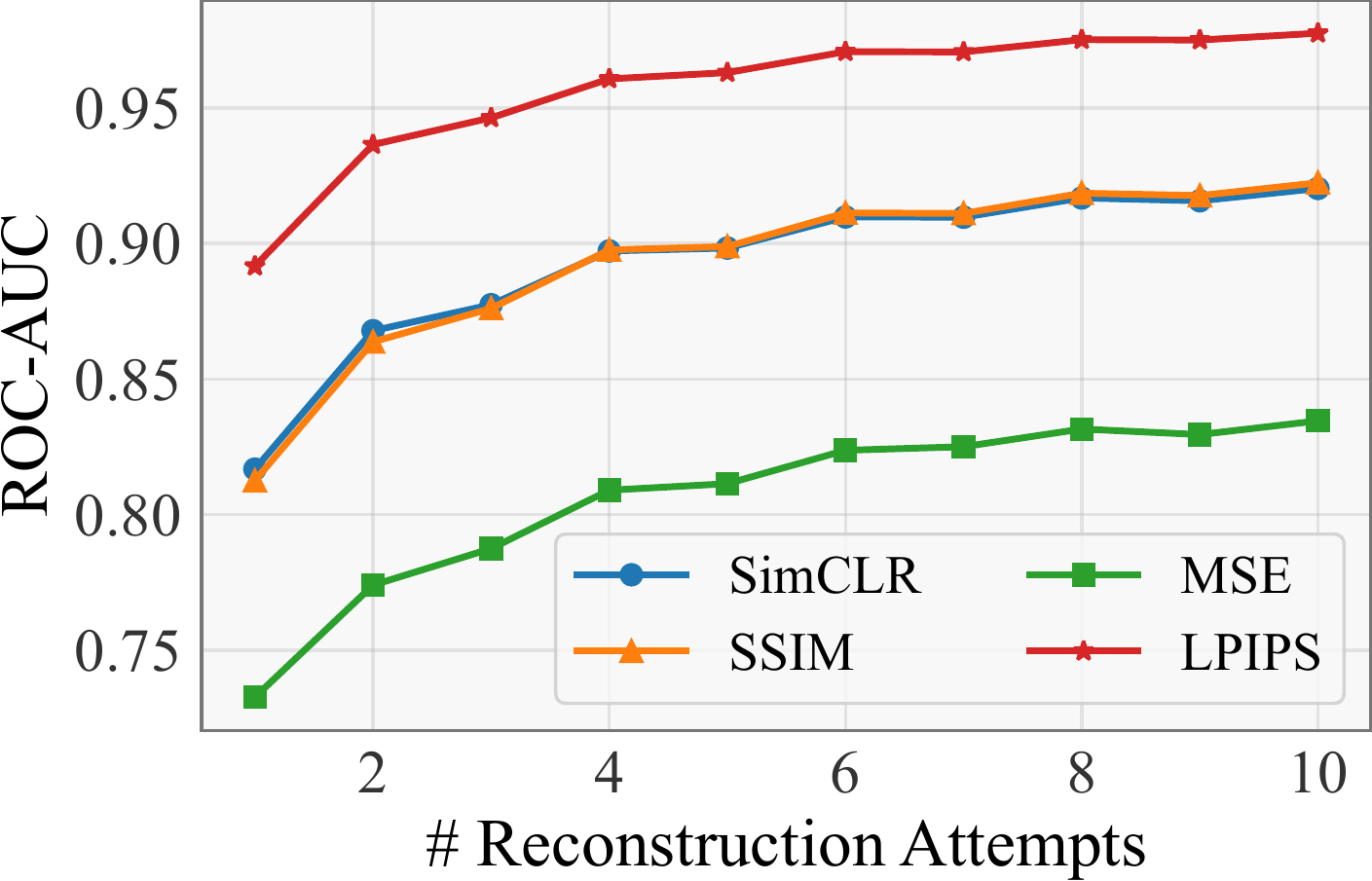}
\caption{KMNIST vs. MNIST}
\end{subfigure}\hfill
\caption{ROC-AUC against number of reconstruction attempts on two pairs of datasets. As the number of reconstruction attempts increases, the OOD detection performance improves regardless of the choice of the reconstruction distance metric.
}
\label{fig:repeats}
\end{figure}

{\bf Number of Reconstruction Attempts per Image.} 
We also study the effect of the number of reconstruction attempts on the performance. Figure~\ref{fig:repeats} shows the ROC-AUC from one attempt to ten attempts per image for two pairs of datasets. In both dataset pairs, increasing the number of attempts almost always improves the ROC-AUC. The improvement is especially significant initially, and saturates at around ten attempts. The improvement is consistent across all four distance metrics, further supporting the effectiveness of \acronym{}'s multiple reconstructions approach.

\begin{table}[h]
 \caption{ROC-AUC performance of our OOD detection framework with an alternative lifting and mapping instantiation -- diffusion and denoising. It shows strong performance, although it is slightly outperformed by our default choice of masking and inpainting.}
 \vskip 0.15in
 \label{tab:denoising}
 \begin{center}
 \begin{small}
 \begin{sc}
 \resizebox{\linewidth}{!}{%
 \begin{tabular}{lcccr}
 \toprule
 Mapping & CIFAR10 & CIFAR10 & FashionMNIST  \\
Method& vs. CIFAR100 & vs. SVHN & vs. MNIST\\
 \midrule
Denoising & 0.600 & 0.976 & 0.941\\
Inpainting & \textbf{0.607} & \textbf{0.992} & \textbf{0.992}\\
 \bottomrule
 \end{tabular}}
 \end{sc}
 \end{small}
 \end{center}
 \vskip -0.1in
 \end{table}

{\bf Alternative Way of Lifting and Mapping.} Alternatively, we can lift an image by diffusion, and map it by denoising. Table~\ref{tab:denoising} shows the performance of our OOD detection framework under this instantiation on three dataset pairs. In our experiments, we add noise to step $t=500$ in each attempt (where $T=1000$), as it generally yields good results. Similar to our inpainting setting, we perform 10 attempts per image, and use the median reconstruction error under LPIPS as the OOD score. We observe that diffusion/denoising is also competitive, although it is slightly outperformed by masking/inpainting. This indicates that our framework is generally applicable in OOD detection, and supports various promising alternative instantiations. 
\section{Discussion and Conclusion}

One limitation of the vanilla diffusion model is that the denoising process involves many iterations and is thus slow. Consequently, like many DM-based algorithms in other applications~\cite{meng2021sdedit,lugmayr2022repaint}, \acronym{} is hard to be applied to real-time OOD detection at the current stage. Recently, there has been a popular line of work on speeding up diffusion models without retraining. For example, \citet{nichol2021improved} re-scales the noise schedule to skip sampling steps, \citet{liu2022pseudo} proposes pseudo numerical methods for diffusion models, and \citet{watson2022learning} optimizes fast samplers that enable sampling with only 10-20 steps. This opens up a promising direction for future work to integrate these methods into \acronym{}.   

In conclusion, we leverage the diffusion model's manifold mapping ability, and propose a method -- Lift, Map, Detect (LMD) -- for unsupervised OOD detection. We show that it is competitive and versatile through our experiments.

\section{Acknowledgement}
This research is supported by grants from DARPA AIE program, Geometries of Learning (HR00112290078), the Natural Sciences and Engineering Research Council of Canada (NSERC) (567916), the National Science Foundation NSF (IIS-2107161, III1526012, IIS-1149882, and IIS-1724282), and the Cornell Center for Materials Research with funding from the NSF MRSEC program (DMR-1719875).
\bibliography{references}
\bibliographystyle{icml2023}

\end{document}